\def\year{2022}\relax
\documentclass[letterpaper]{article} 
\usepackage{aaai22}  
\usepackage{times}  
\usepackage{helvet}  
\usepackage{courier}  
\usepackage[hyphens]{url}  
\usepackage{graphicx} 
\usepackage{natbib}  
\usepackage{caption} 
\DeclareCaptionStyle{ruled}{labelfont=normalfont,labelsep=colon,strut=off} 
\usepackage{xr}
\usepackage{epsfig}
\usepackage{graphicx,comment}
\usepackage{amsmath}
\usepackage{indentfirst}
\usepackage{algorithm}
\usepackage{algpseudocode}
\usepackage{amssymb}
\usepackage{verbatim}
\usepackage{multirow}
\usepackage{booktabs}
\usepackage{makecell}
\usepackage{longtable}
\usepackage{soul}
\usepackage{dirtytalk}
\usepackage{caption}
\usepackage{natbib}
\usepackage{arydshln}
\usepackage{xcolor}
\usepackage{float}
\usepackage{eso-pic}
\usepackage{xspace}
\usepackage{kotex}
\usepackage{bm, subfigure}
\usepackage[titletoc]{appendix}
\usepackage[resetlabels]{multibib}
\newcites{sec}{References}
\usepackage{newfloat}
\usepackage{listings}
\floatstyle{ruled}
\newfloat{listing}{tb}{lst}{}
\floatname{listing}{Listing}
\newcommand\tab[1][0.25cm]{\hspace*{#1}}

\newcommand{\SystemName}{ADD}

\definecolor{mediumcandyapplered}{rgb}{0.89, 0.02, 0.17}
\definecolor{mordantred19}{rgb}{0.68, 0.05, 0.0}
\definecolor{oxfordblue}{rgb}{0.0, 0.13, 0.28}

\usepackage{amsmath}
\DeclareMathOperator*{\argmax}{arg\,max}

\makeatletter
\def\adl@drawiv#1#2#3{%
        \hskip.5\tabcolsep
        \xleaders#3{#2.5\@tempdimb #1{1}#2.5\@tempdimb}%
                #2\z@ plus1fil minus1fil\relax
        \hskip.5\tabcolsep}
\newcommand{\cdashlinelr}[1]{%
  \noalign{\vskip\aboverulesep
          \global\let\@dashdrawstore\adl@draw
          \global\let\adl@draw\adl@drawiv}
  \cdashline{#1}
  \noalign{\global\let\adl@draw\@dashdrawstore
          \vskip\belowrulesep}}
\makeatother

\usepackage[breaklinks=false, bookmarks=false, backref=page, colorlinks=true, linkcolor=red, citecolor=black]{hyperref}

\renewcommand*{\backref}[1]{}
\renewcommand*{\backrefalt}[4]{%
    \ifcase #1  (Not cited.)%
    \or         (Cited on page~#2.)%
    \else       (Cited on pages~#2.)%
    \fi}
\newcounter{alphasect}
\def\alphainsection{0}

\let\oldsection=\section
\def\section{%
  \ifnum\alphainsection=1%
    \addtocounter{alphasect}{1}
  \fi%
\oldsection}%

\renewcommand\thesection{%
  \ifnum\alphainsection=1%
    \Alph{alphasect}
  \else%
    \arabic{section}
  \fi%
}%

\newenvironment{alphasection}{%
  \ifnum\alphainsection=1%
    \errhelp={Let other blocks end at the beginning of the next block.}
    \errmessage{Nested Alpha section not allowed}
  \fi%
  \setcounter{alphasect}{0}
  \def\alphainsection{1}
}{%
  \setcounter{alphasect}{0}
  \def\alphainsection{0}
}%
\newcommand\Tstrut{\rule{0pt}{2.6ex}}       
\newcommand\Bstrut{\rule[-1.1ex]{0pt}{0pt}} 

\title{ADD: Frequency Attention and Multi-View based Knowledge Distillation to Detect Low-Quality Compressed Deepfake Images}
\author {
    Binh M. Le,\textsuperscript{\rm 1}
    Simon S. Woo, \textsuperscript{\rm 2}
}
\affiliations {
    \textsuperscript{\rm 1} Department of Computer Science and Engineering, Sungkyunkwan University, South Korea\\
    \textsuperscript{\rm 2} Department of Applied Data Science, Sungkyunkwan University, South Korea\\
    bmle@g.skku.edu, swoo@g.skku.edu
}

\begin{document}

\maketitle
\begin{abstract}
Despite significant advancements of deep learning-based forgery detectors for distinguishing manipulated deepfake images, most detection approaches suffer from moderate to significant performance degradation with low-quality compressed deepfake images.Because of the limited information in low-quality images, detecting low-quality deepfake remains an important challenge. In this work, we apply frequency domain learning and optimal transport theory in knowledge distillation (KD) to specifically improve the detection of low-quality compressed deepfake images. We explore transfer learning capability in KD to enable a student network to learn discriminative features from low-quality images effectively. In particular, we propose the Attention-based Deepfake detection Distiller~(\SystemName), which consists of two novel distillations: 1) frequency attention distillation that effectively retrieves the removed high-frequency components in the student network, and 2) multi-view attention distillation that creates multiple attention vectors by slicing the teacher’s and student’s tensors under different views to transfer the teacher tensor’s distribution to the student more efficiently. Our extensive experimental results demonstrate that our approach outperforms state-of-the-art baselines in detecting low-quality compressed deepfake images.
\end{abstract}
\section{Introduction}
Recently, facial manipulation techniques using deep learning methods such as deepfakes 
have drawn considerable attention ~\cite{rossler2019faceforensics++, pidhorskyi2020adversarial, richardson2020encoding, nitzan2020face}. Moreover, deepfakes have become more realistic and sophisticated, making it difficult to be distinguished by human eyes \cite{siarohin2020first}. And it has become much easier to generate such realistic deepfakes than before. Hence, such advancements and convenience enable even novices to easily create highly realistic fake faces for simple entertainment. However, these fake images raise serious security, privacy, and social concerns, as they can be abused for malicious purposes, such as impersonation~\cite{fraudst}, revenge pornography~\cite{ cole_2018}, and fake news propagation~\cite{quandt2019fake}.  

To address such problems arising from deepfakes, there have been immense research efforts in developing effective deepfake detectors \cite{dzanic2019fourier, rossler2019faceforensics++, wang2019fakespotter, li2018exposing, khayatkhoei2020spatial, zhang2019detecting}. 
Most approaches utilize the deep learning-based approaches, where generally they perform well if there are a large amount of high-resolution training data. However, the performances of these approaches drop dramatically (by up to $18\%$~\cite{dzanic2019fourier, rossler2019faceforensics++}) for compressed low-resolution images due to lack of available pixel information to sufficiently distinguish fake images from real ones. In other words, because of the compression, subtle differences and artifacts such as sharp edges in hairs and lips that can be possibly leveraged for differentiating deepfakes can also be removed. Therefore, there still remains an important challenge to effectively detect low-quality compressed deepfakes, which frequently occur on social media and mobile platforms in bandwidth-challenging and storage-limited environments.   

\begin{figure}[t!]
\centering
\includegraphics[width=3.1 in]{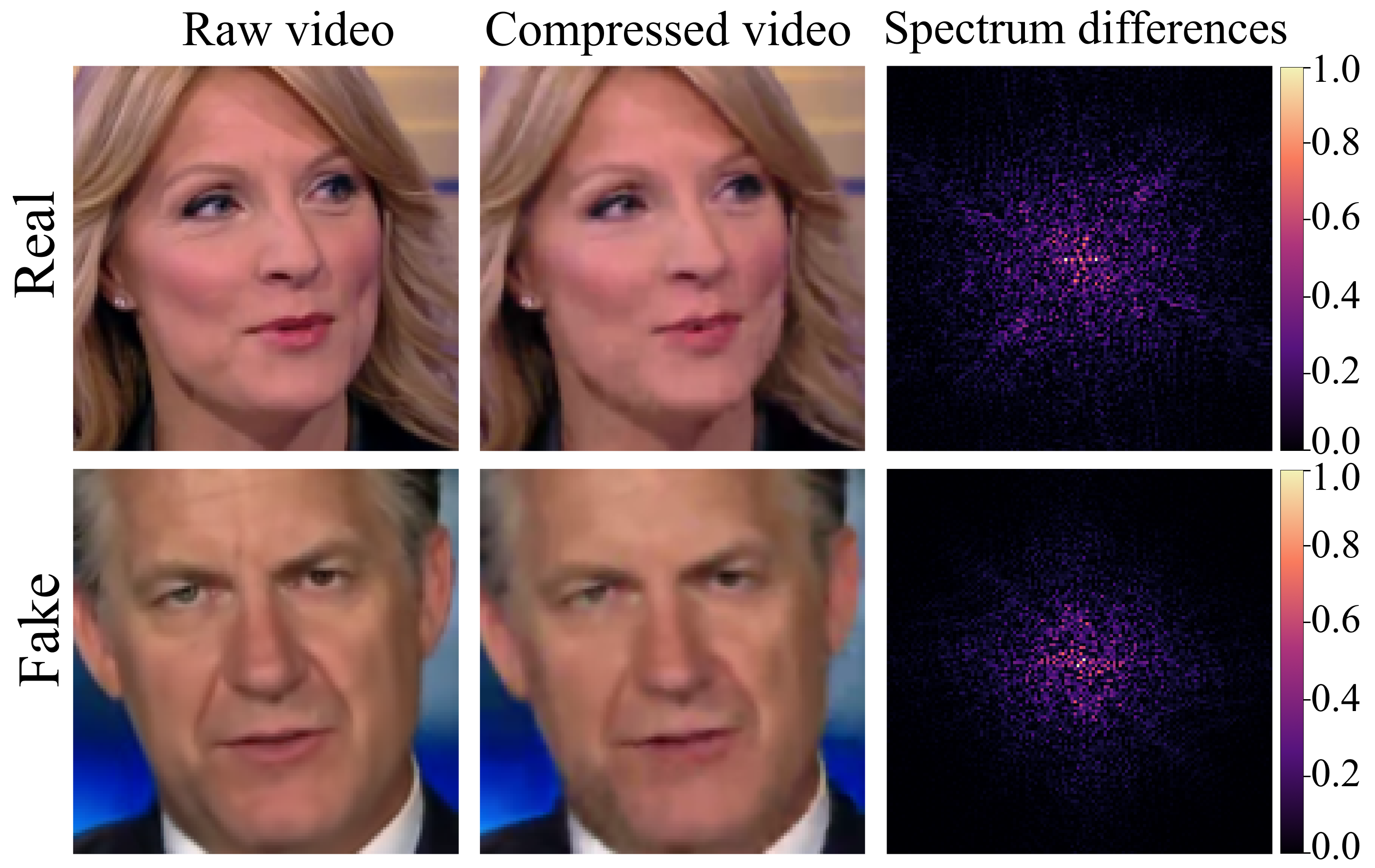}
\caption{Degradation of high-frequency components after compression of real and fake videos. Sample fake face frames are taken from the NeuralTextures dataset in FaceForensics++~\cite{rossler2019faceforensics++}. \textbf{Left column}: Sample faces from raw videos. \textbf{Middle column}: Sample faces from c40-compressed videos. \textbf{Right column}: Normalized spectrum differences in the frequency domain after applying Discrete Fourier Transform (DFT) to raw and compressed images. The concentrated differences at the center are the highest frequency components.} 
\label{fig:raw_vs_compress}
\end{figure}

In this work, we propose the Attention-based Deepfake detection Distiller (ADD). Our primary goal is to detect low-quality (LQ) deepfakes, which are less explored in most previous studies but plays a pivotal role in real-world scenarios. First, we assume there are high-quality (HQ) images are readily available, similar to the settings in other studies~\cite{ rossler2019faceforensics++, wang2019fakespotter, li2018exposing, khayatkhoei2020spatial, zhang2019detecting,dzanic2019fourier}. And, we use knowledge distillation (KD) as an overarching backbone architecture to detect low-quality deepfakes. While most of the existing knowledge distillation methods aim to reduce the student size for model compression applications or improve the performance of  lightweight deep learning models ~\cite{hinton2015distilling, tian2019contrastive, huang2017like, passalis2018learning}, we hypothesize that a student can learn lost distinctive features of low-quality compressed images from a teacher that is pre-trained on high-quality images for deepfake detection.   
We first lay out the following two major challenges associated with detecting the LQ compressed deepfakes, and provide the intuitions of our approaches to overcome these issues:

\textbf{1) Loss of high-frequency information.} 
As discussed, while lossy image compression algorithms make changes visually unnoticeable to humans, they can significantly reduce DNNs' deepfake detection capability  by removing the fine-grained artifacts in high-frequency components. To investigate this phenomenon more concretely, we revisit Frequency Principal (F-Principal) \cite{xu2019training}, which describes the learning behavior of general DNNs in the frequency domain. F-Principal states that general DNNs tend to learn dominant low-frequency components first and then capture high-frequency components during the training process \cite{xu2019training}. For example, to illustrate this issue, Fig.~\ref{fig:raw_vs_compress} is provided to indicates that most of the lost information during compression is from high-frequency components.
As a consequence, general DNNs shift their attention in later training epochs to high-frequency components, which now represent intrinsic characteristics of objects in each individual image rather than discriminative features. This learning process increases the variance of DNNs' decision boundaries and induces overfitting, thereby degrading the detection performance. A trivial approach to tackle the overfitting is applying the early stopping method \cite{morgan1989generalization}; however, fine-grained artifacts of deepfakes can be subsequently omitted, especially when they are highly compressed.
To overcome this issue, we propose the novel frequency attention distiller, which guides the student to effectively recover the removed high-frequency components in low-quality compressed images from the teacher during training.

\textbf{2) Loss of correlated information}. In addition, under heavy compression, crucial features and pixel correlations that not only capture the intra-class variations, but also characterize the inter-class differences are also degraded. In particular, these correlations are essential for CNNs' ability to learn the features at the local filters, but they are significantly removed in the compressed input images.

\begin{figure*}[t!]
\centering
\includegraphics[width=0.8\linewidth]{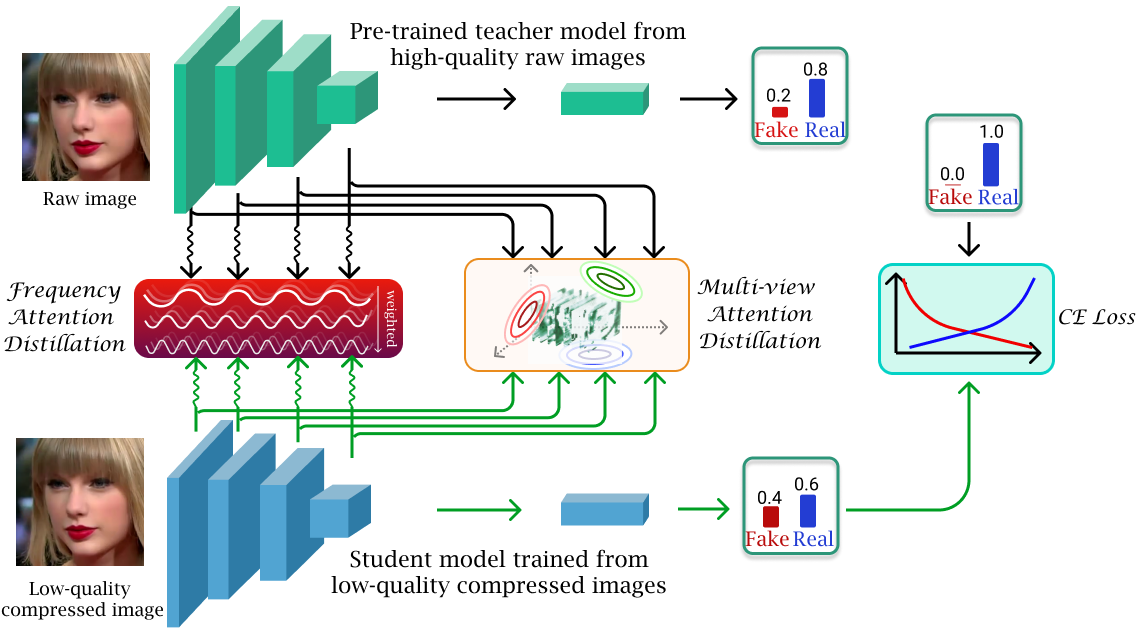}
\caption{Illustration of our proposed Attention-based Deepfake detection Distiller~(\SystemName)~distillation framework. First, a low-quality compressed image and its corresponding raw image are used as an input to the student and pre-trained teacher model, respectively. The student model is trained with two novel distillers: 1) frequency attention distiller and 2) multi-view attention distiller. The frequency attention distiller creates a weighted loss, which focuses more on the degraded high-frequency components. The multi-view attention distiller slices student's and teacher's tensors by different random views to spawn multiple attention vectors. Green arrows indicate the flows of gradient decent updates to train the student's parameters.
}
\label{fig:overall_arc}
\end{figure*}

Recent studies~\cite{wang2018non, hu2018relation} have empirically demonstrated that training DNNs that are able to capture this correlated information can successfully improve their performances. Therefore, in this work, we focus on improving the lost correlations by proposing a novel multi-view attention, inspired by the work of Bonneel \emph{et al.} \cite{bonneel2015sliced}, and contrastive distillation~\cite{tian2019contrastive}.
The element-wise discrepancy between the teacher's and student's tensors that ignores the relationship within local regions of pixels, or channel-wise attention that only considers a single dimension of backbone features. On the other hand, our proposed method ensures that our model attends to output tensors from multiple views (slices) using Sliced Wasserstein distance (SWD)~\cite{bonneel2015sliced}. Therefore, our multi-view attention distiller guides the student to mimic its teacher more efficiently through a geometrically meaningful metric based on SWD. In summary, we present our overall Attention-based Deepfake detection Distiller (ADD), which consists of two novel distillations 
(See Fig.~\ref{fig:overall_arc}): 1) frequency attention distillation that effectively retrieves the removed high-frequency components in the student network, and 2) 
multi-view attention distillation that creates multiple attention vectors by slicing the teacher's and student's tensors under different views to transfer the teacher tensor's distribution to the student more efficiently. 

Our contributions are summarized as follows:
\begin{itemize}
    \item We propose the novel \emph{frequency attention distillation}, which effectively enables the student to retrieve high-frequency information from the teacher.
    
    \item We develop the novel \emph{multi-view attention distillation} with contrastive distillation for the student to efficiently mimic the teacher while maintaining pixel correlations from the teacher to the student through SWD. 

    \item We demonstrate that our approach outperforms well-known baselines, including attention-based distillation methods, on different low-quality compressed deepfake datasets.
\end{itemize}

\section{Related Work}
\textbf{Deepfake detection. } Deepfake detection has recently drawn significant attention, as it is related to protecting personal privacy. Therefore, there has been a large number of research works to identify such deepfakes \cite{rossler2019faceforensics++, li2020face,li2019faceshifter,jeon2020t, rahmouni2017distinguishing, wang2019fakespotter, li2018exposing}. Li \emph{et al.} \cite{li2020face} tried to expose the blending boundaries of generated faces and showed the effectiveness of their method, when applied for unseen face manipulation techniques. Self-training with L2-starting point regularization was introduced by Jeon \emph{et al.} \cite{jeon2020t} to detect newly generated images. However, the majority of prior works are limited to high-quality (HQ) synthetic images, which are rather straightforward to detect by constructing binary classifiers with a large amount of HQ images.

\textbf{Knowledge distillation (KD). } Firstly introduced by Hinton \emph{et al.} \cite{hinton2015distilling}, KD is a training technique that transfers acquired knowledge from a pre-trained teacher model to a student model for model compression applications. However, many existing works \cite{yim2017gift, tian2019contrastive, huang2017like, passalis2018learning} applied different types of distillation methods to conventional datasets, \emph{e.g.}, ImageNet, PASCAL VOC 2007, and CIFAR100, but not for deepfake datasets. 
On the other hand, Zhu \emph{et al.} \cite{zhu2019low} used FitNets \cite{romero2014fitnets} to train a student model that is able to detect low-resolution images, which is similar to our method in that the teacher and the student learn to detect high and low-quality images, respectively. However, their approach coerces the student to mimic the penultimate layer's distribution from the teacher, while it does not possess rich features at the lower layers.

In order to encourage the student model to mimic the teacher more effectively, Zagoruyko and Komodakis \cite{zagoruyko2016paying} proposed the activation-based attention transfer, similar to FitNets, but their approach achieves better performance by creating spatial attention maps. Our multi-view attention method inherits from this approach but carries more generalization ability by not only exploiting spatial attention (in width and height dimension), but also introducing attention features from random dimensions using Radon transform \cite{helgason2010integral}. Thus, our approach pushes the student's backbone features closer to the teacher's.

In addition, inspired by InfoNCE loss \cite{oord2018representation}, Tian \emph{et al.} \cite{tian2019contrastive} proposed contrastive representation distillation (CRD), which formulates the contrastive learning framework and motivates the student network to drive samples from positive pairs closer, and push away those from negative pairs. Although CRD achieves superior performance to those of previous approaches, it requires a large memory buffer to save embedding features of each sample. This is restrictive when training size and embedding space become larger.  Instead, we directly sample positive and negative images in the same mini-batch and apply the contrastive loss to embedded features, similar to the Siamese network \cite{bromley1994signature}.

\textbf{Frequency domain learning. }In the field of media forensics, several approaches~\cite{jiang2020focal, khayatkhoei2020spatial, dzanic2019fourier} showed that discrepancies of high-frequency's Fourier spectrum are effective clues to distinguish CNN-based generated  images. Frank \emph{et al.} \cite{frank2020leveraging} and  and Zhang \emph{et al.} \cite{ zhang2019detecting} utilized the checkerboard artifacts \cite{odena2016deconvolution} of the frequency spectrum caused by up-sampling components of generative neural networks (GAN) as effective features in detecting GAN-based fake images. Nevertheless, their detection performances were greatly degraded when the training synthesized images are compressed, becoming low-quality. Quian \emph{et al.} proposed an effective frequency-based forgery detection method, named $F^3Net$, which decomposes an input image to many frequency components, collaborating with local frequency statistics on a two-streams network. The $F^3Net$, however, doubles the number of parameters from its backbone.

\textbf{Wasserstein distance.} Induced by the optimal transport theory, Wasserstein distance (WD)~\cite{villani2008optimal}, and its variations have been explored in training DNNs to learn a particular distribution thanks to Wasserstein's underlying geometrically meaningful distance property. In fact, WD-based applications cover a wide range of fields, such as to improve generative models \cite{arjovsky2017wasserstein, deshpande2018generative}, learn the distribution of latent space in autoencoders \cite{kolouri2018sliced, xu2020learning}, and match features in domain adaptation tasks \cite{lee2019sliced}.In this work, we utilize the Wasserstein metric to provide the student geometrically meaningful guidance to efficiently mimic the teacher's tensor distribution. Thus, the student can learn the true tensor distribution, even though its input features are partially degraded through high compression.
\section{Our Approach} \label{methodology}
Our Attention-based Deepfake detection Distiller (ADD) is consisted of the following two novel distillations 
(See Fig.~\ref{fig:overall_arc}): 1) frequency attention distillation and 2)  multi-view attention distillation.
\subsection{Frequency Attention Distillation}
\begin{figure}[t!]
\centering
\includegraphics[width=3.1in]{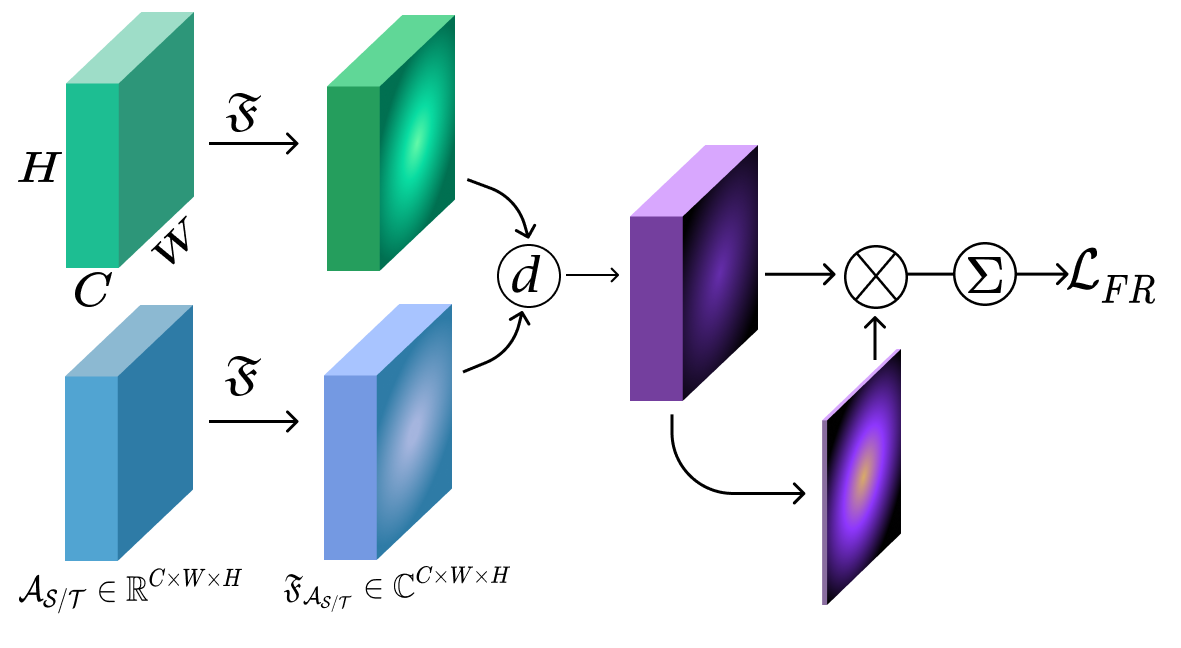}
\caption{Illustration of our frequency attention distiller. The $\mathfrak{F}$ function is applied to each channel of the input tensor. Distance metric $d$ calculates the discrepancy of the corresponding coefficients of each frequency component from the teacher and the student. Finally, the attention map is obtained by averaging the element-wise differences across channels.}
\label{fig:freq_att}
\end{figure}

Let $f_{\mathcal{S}}$ and $f_{\mathcal{T}}$ be the student and the pre-trained teacher network. By forwarding a low-quality compressed input image and its corresponding raw image through $f_{\mathcal{S}}$ and $f_{\mathcal{T}}$, respectively, we obtain features $\mathcal{A}_{\mathcal{S}}$ and $\mathcal{A}_{\mathcal{T}} \in \mathbb{R}^{C \times W \times H} $ from its backbone network, which have $C$ channels, the width of $W$, and the height of $H$.
To create frequency representations, Discrete Fourier Transform (DFT) $\mathfrak{F}: \mathbb{R}^{C \times W \times H} \to \mathbb{C}^{C \times W \times H}$ is applied to each channel as follows:
\begin{equation}
\mathfrak{F}_{\mathcal{A} _{\mathcal{S}/ \mathcal{T}}}(c,u,v)=\displaystyle\sum_{x=1}^{W} \sum_{y=1}^{H} \mathcal{A} _{\mathcal{S}/ \mathcal{T}}(c, x, y)\cdot e^{-i2\pi(\frac{ux}{W}+ \frac{vy}{H})},
\label{eqn:fft}
\end{equation}
where $c, x$ and $y$ denote the $c_{th}, x_{th}$ and $y_{th}$ slice in the channel, the width and height dimension of 
$\mathcal{A}  _{\mathcal{S}}$ and $\mathcal{A}  _{\mathcal{T}}$, respectively. Here, for convenience, we use the notation $\mathfrak{F}_{\mathcal{A} _{\mathcal{S}/ \mathcal{T}}}$ to denote that the function is independently applied for both student's and teacher's backbone features. Then, the value at $(u,v)$ on each single feature-map $\mathfrak{F}_{\mathcal{A} _{\mathcal{S}/ \mathcal{T}}}(c,:,:)$ indicates the coefficient of a basic frequency component. The difference between a pair of corresponding coefficients from the teacher and the student represents the \say{absence} of that student's frequency component. Next, let $d: \mathbb{C}^2 \to \mathbb{R}^{+}$ be a metric that assesses the distance between two input complex numbers and supports stochastic gradient descent.
Then, the frequency loss between the teacher and student can be defined as follows:
\begin{equation}
\mathcal{L}_{\mathcal{FR}} = \displaystyle\sum_{c=1}^{C}\sum_{u=1}^{W} \sum_{v=1}^{H} w  (u, v) \cdot d\big(\mathfrak{F}_{\mathcal{A} _{\mathcal{S}}}(c,u,v), \mathfrak{F}_{\mathcal{A} _{\mathcal{T}}}(c,u,v)\big),
\label{eqn:1}
\end{equation}
where $w  (u, v)$ is an attention weight at $(u, v)$. In this work, we utilize the exponential of the difference across channels between the teacher and student as the weight in the following way:
\begin{equation}
w (u, v) = exp\big(\gamma_{\mathcal{FR}} \cdot \frac{1}{C}\displaystyle\sum_{c=1}^{C} d(\mathfrak{F}_{\mathcal{A} _{\mathcal{S}}}(c,u,v), \mathfrak{F}_{\mathcal{A}_{\mathcal{T}}}(c,u,v))\big),
\label{eqn:2}
\end{equation}
where $\gamma_{\mathcal{FR}}$ is a positive hyper-parameter that governs the exponential cumulative loss, as the student's removed frequency increases. This design of attention weights ensures that the model focuses more on the losing high-frequency, and makes Eq. \ref{eqn:1} partly similar to  \emph{focal loss} \cite{lin2017focal}. Figure \textcolor{red}{\ref{fig:freq_att}} visually illustrates our frequency loss.\\

\subsection{Multi-view Attention Distillation}

\textbf{Sliced Wasserstein distance. } The \emph{p}-Wasserstein distance between two probability measures $\mu$ and $\nu$ \cite{villani2008optimal} with their corresponding probability density functions $P_{\mu}$ and $P_{\nu}$ in a probability space $(\Omega,\mathcal{P}(\Omega) )$ and $\Omega \subset \mathbb{R}^d$, is defined as follows:
 \begin{equation}
 \label{eqn:4}
    W_p(P_{\mu},P_{\nu}) = \Big(\inf_{\pi \in \Pi(\mu, \nu)}  \int_{\Omega \times \Omega}\psi(x,y)^p d\pi(x,y) \Big)^{1/p},
\end{equation}
 where $\Pi(\mu, \nu)$ is a set of all transportation plans $\pi$, which has the marginal densities $P_{\mu}$ and $P_{\nu}$, respectively, and $\psi:\Omega \times \Omega \to \mathbb{R}^+$ is a transportation cost function. Equation \textcolor{red}{\ref{eqn:4}} searches for an optimal transportation plan between $\mu$ and $\nu$, which is also known as Kantorovitch formulation \cite{kantorovitch1958translocation}. In the case of one-dimensional probability space, \emph{i.e.}, $\Omega \subset \mathbb{R}$, the closed-form solution of the \emph{p}-Wasserstein distance is:
 \begin{equation}
    W_p(P_{\mu},P_{\nu}) =  \Big(\int_{0}^{1} \psi \big(F_{\mu}^{-1}(\kappa), F_{\nu}^{-1}(\kappa)\big)^p \,d\kappa\Big) ^{1/p},
 \end{equation}
 where $F_{\mu}$ and $F_{\nu}$ are the cumulative distribution functions of $P_{\mu}$ and $P_{\nu}$, respectively.\par
 
A variation of Wasserstein distance, inspired by the above closed-form solution, is Sliced Wasserstein distance (SWD) that deploys multiple projections from a high dimensional distribution to various one-dimensional marginal distributions and calculates the optimal transportation cost for each projection. In order to construct these one-dimensional marginal distributions, we use the Radon transform~\cite{helgason2010integral}, which is defined as follows:
\begin{equation}
    \mathcal{R}_{P_{\mu}}(t,\theta)=\int_{\Omega} \mu (x) \delta (t-\langle \theta, x \rangle) dx, \forall \theta \in \mathcal{S}^{d-1}, \forall t \in \mathbb{R},
\end{equation}
where $\delta$ denotes the Diract delta function, $\langle \cdot,\cdot \rangle$ is the Euclidean inner-product, and $\mathcal{S}^{d-1} \subset \mathbb{R}^d$ is the \emph{d}-dimensional unit sphere. Thus, we denote $\mathcal{R}_{\theta} \mu$ as a 1-D marginal distribution of $\mu$ under the projection on $\theta$.
The Sliced 1-Wasserstein distance is defined as follows:
\begin{equation}
\label{eqn:6}
    SW_{1}(P_{\mu}, P_{\nu}) = \int_{{S}^{d-1}}W_1(\mathcal{R}_{\theta }\mu, \mathcal{R}_{\theta  }\nu)d\theta.
\end{equation} \par
Now, we can calculate the Sliced Wasserstein distance by optimizing a series of 1-D transportation problems, which have the closed-form solution that can be computed in $\mathcal{O}(Nlog(N))$ \cite{rabin2011wasserstein}. In particular, by sorting $\mathcal{R}_{\theta}\mu$ and $\mathcal{R}_{\theta}\nu$ in ascending order using two permutation operators $\tau_{1}$ and $\tau_{2}$, respectively, the $SWD$ can be approximated as follows:
\begin{equation}
    SWD(P_{\mu}, P_{\nu}) \approx \sum_{k=1}^{K} \sum_{i=1}^{N}\psi(\mathcal{R}_{\theta_k}\mu_{\tau_{1}[i]}, \mathcal{R}_{\theta_k}\nu_{\tau_{2}[i]}),
\label{eqn:7}
\end{equation}
where $K$ is the number of uniform random samples $\theta$ using Monte Carlo method to approximate the integration of $\theta$ over the unit sphere $\mathcal{S}^{d-1}$ in Eq. \textcolor{red}{\ref{eqn:6}}.

\textbf{Multi-view attention distillation. } Let $P_{\mathcal{A} }$ be the square of $\mathcal{A}$ after being normalized by the Frobenius norm, \emph{i.e.}, $P_{\mathcal{A} } = \frac{\mathcal{A}^{\circ 2} }{\Vert \mathcal{A} \Vert_F^2} $, where $\circ$ denotes the Hadamard power \cite{bocci2016hadamard}. Consequently, we are now able to consider $P_{\mathcal{A} }$ as a discrete probability density function over $\Omega=\mathbb{R}^{C\times W \times H} \subset \mathbb{R}^3$, where $P_{\mathcal{A} }(c,x,y)$ indicates the density value at the slice $c_{th}, x_{th}$ and $y_{th}$ of the channel, the width and height dimension, respectively. To avoid replicating the element-wise differences,  we additionally need to bin the projected vectors into $g$ groups before applying distillation. One important property of our multi-view attention is that different values of $\theta$ provide different attention views (slices) of $\mathcal{A} _{\mathcal{S}}$ and $\mathcal{A} _{\mathcal{T}}$. For instance, with $\theta=(1,0,0)$, we achieve the channel-wise attention that was introduced by Chen \emph{et al.} \cite{chen2017sca}. Or, we can produce an attention vector in the width and height dimension, when $\theta$ becomes close to $(0,1,0)$ and $(0,0,1)$, respectively. With this general property, a student can pay full attention to its teacher's tensor distribution instead of some pre-defined constant attention views.

\begin{figure}[t!]
\centering
\includegraphics[width=3.1in]{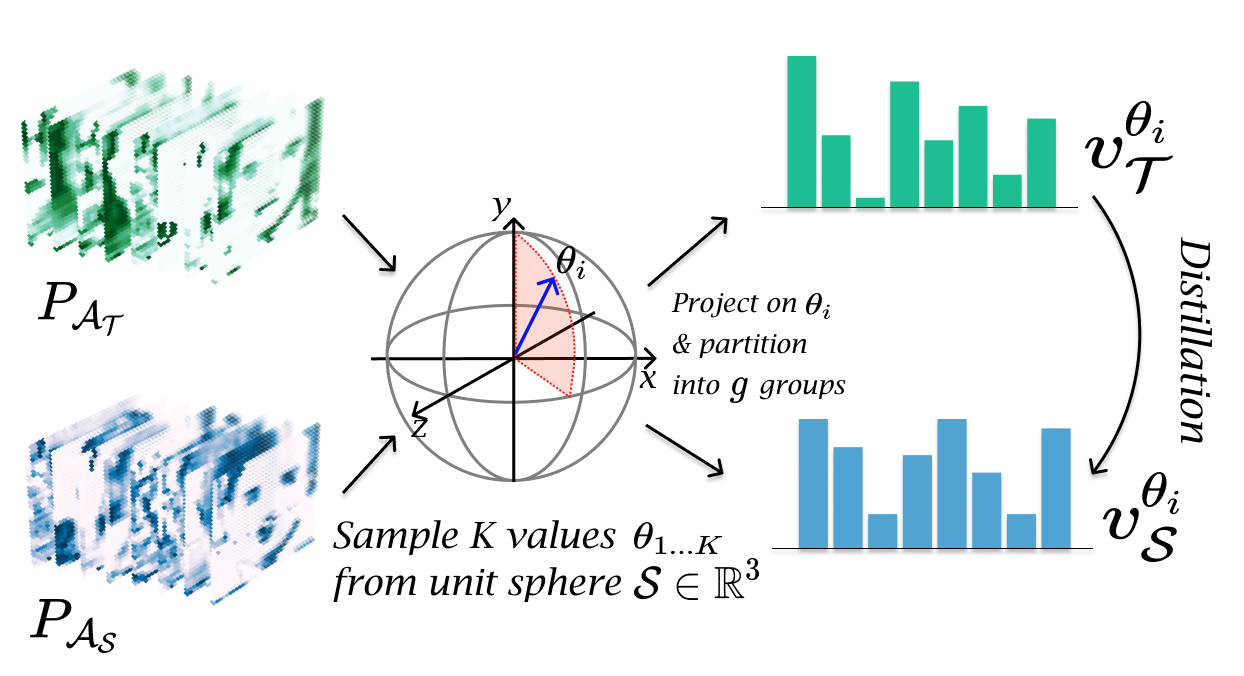}
\caption{Detailed illustration of our multi-view attention distillation. Two backbone features of the teacher (top) and the student (bottom). After normalization, obtained features are projected on a random sample $\theta_i$, then two attention vectors, $v_{\mathcal{T}}^{\theta_i}$ and $v_{\mathcal{S}}^{\theta_i}$, are obtained by sorting the projection images and binning them into $g $ groups. Multiple values of $\theta$ provides us multiple attention views on the two tensors.}
\label{fig:multi_view}
\end{figure}

Figure \textcolor{red}{\ref{fig:multi_view}} pictorially illustrates our overall multi-view attention distillation, and we summarize our multi-view attention in Algorithm \ref{alg:multiview_attention} in the supplementary materials.
In order to encourage the semantic similarity of samples' representation from the same class and discourage that of those from different classes, we further apply the contrastive loss for each instance, which inspired by the CRD distillation framework of Tian \emph{et al.} \cite{tian2019contrastive} . Thus, our overall multi-view attention loss is defined as follows:
\begin{equation}
\begin{split}
    \mathcal{L}_{\mathcal{MV}}   = & \gamma_{\mathcal{MV}} \times  SWD(P_{\mathcal{A} _{\mathcal{S}}, P_{\mathcal{A} _{\mathcal{T}}}}) + \\
    & \eta_{\mathcal{MV}} \times \big[ SWD(P_{\mathcal{A} _{\mathcal{S}}}, P_{\mathcal{A}^{+}_{\mathcal{T}}}) + \\
     & \max(\Delta - SWD\big(P_{\mathcal{A} _{\mathcal{S}}}, P_{\mathcal{A}^{-}_{\mathcal{T}}}), 0\big) \big],
\end{split}
\label{eqn:8}
\end{equation}
where $\mathcal{A}^{+}_{\mathcal{T}}$ and $\mathcal{A}^{-}_{\mathcal{T}}$ are the random instance's representation that belong to the same and the opposite class of ${A}  _{\mathcal{S}}$ at the teacher, respectively. And $\Delta$ is a margin that manages the discrepancy of negative pairs and $\gamma_{\mathcal{MV}}$, and $\eta_{\mathcal{MV}}$ are scaling hyper-parameters.

\subsection{Overall Loss Function }

The overall distillation loss in our KD framework is formulated as follows:
\begin{equation}
 \mathcal{L}_{Distill}(A_{\mathcal{S}}, A_{\mathcal{T}}) =  \underbrace{\alpha \cdot \mathcal{L}_{\mathcal{FR}}}_\text{frequency attention} + \underbrace{\beta \cdot \mathcal{L}_{\mathcal{MV}}}_\text{multi-view attention} ,
 \label{eqn:10}
\end{equation}
where $\alpha$ and $\beta$ are hyper-parameters to balance the contribution of  frequency attention distiller and multi-view attention distiller, respectively. Our attention loss is parameter-free and is independent from model architecture design, and it can be directly added to any detector model's conventional loss (\emph{e.g.}, cross-entropy loss).  Also, the frequency attention requires computational complexity in $\mathcal{O} \big( CWH \cdot (log(W)+log(H))  \big)$  for one backbone feature, where $\mathcal{O} \big(WH \cdot (log(W)+log(H))  \big)$ is the complexity of 2-D Fast Fourier Transform applied for one channel. On the other hand, the average-case complexity of multi-view attention is $\mathcal{O} \big(KN \cdot log(N)\big)$, where $\mathcal{O}\big(N \cdot log(N)\big)$ is the complexity of 1-D closed-form solution as mentioned above, $K$ is the number of random samples $\theta$, and $N$ is the number of elements in one backbone feature, \emph{i.e.}, $N=CWH$. Our end-to-end Attention-based Deepfake detection Distiller (ADD) pipeline is presented in Fig.~\ref{fig:overall_arc}.

\section{Experiment}
\subsection{Datasets}
Our proposed method is evaluated on five different popular deepfake benchmark datasets: NeuralTextures \cite{thies2019deferred}, Deepfakes \cite{deepfakes}, Face2Face \cite{thies2016face2face}, FaceSwap \cite{faceswap}, and FaceShifter \cite{li2019faceshifter}. Every dataset has 1,000 videos generated from 1,000 real human face videos by R\"ossler \emph{et al.} \cite{rossler2019faceforensics++}. These videos are compressed into two versions: medium compression (c23) and high compression (c40), using  the H.264 codec with a constant rate quantization parameter of 23, and 40, respectively. Each dataset is randomly divided into training, validation, and test set consisting of 720, 140, and 140 videos, respectively. We randomly select 64 frames from each video and obtain 92,160, 17,920, and 17,920 images for training, validation, and test set, respectively. Then, we utilize the Dblib \cite{king2009dlib} to detect the largest face in every single frame and resize them to a square image of $128 \times 128$ pixels. 

\subsection{Experiment Settings}
In our experiments, we use Adam optimizer \cite{kingma2014adam} with $\beta_1=0.9$, $\beta_2=0.999$ and $\epsilon=10^{-8}$. The learning rate is $2\times 10^{-4}$, which follows one cycle learning rate schedule \cite{smith2019super} with a mini-batch size of 144. In every epoch, the model is validated 10 times to save the best parameters using validation accuracy. Early stopping is applied, when the validation performance does not improve after 10 consecutive times.
We use ResNet50 \cite{he2016deep} as our backbone to implement our proposed distillation framework. 
In Eq. \textcolor{red}{\ref{eqn:1}}, we define $d$ as the square of the modulus of the difference between two complex number, \emph{i.e.}, $d(c_1,c_2)=|c_1-c_2|^2$, which satisfies the properties of a general distance metric: non-negative, symmetric, identity, and triangle inequality. 
The number of binning groups $g$ is equal to a half of the number of channels of $\mathcal{A}_{\mathcal{S}}$. Our hyper-parameters settings $\{\gamma_{\mathcal{FR}}=1,\gamma_{\mathcal{MV}}=100, \eta_{\mathcal{MV}}=50, \Delta=0.012, \alpha=1, \}$ are kept the same, while $\beta$ is  \textbf{fine-tuned} on each dataset in the range of 16 to 23 through the experiments. The experiments are conducted on two TITAN RTX 24GB GPUs with Intel Xeon Gold 6230R CPU @ 2.10GHz.

\subsection{Results}
Our experimental results are presented in Table  \textcolor{red}{\ref{table:results_main}}. We use Accuracy score (ACC) and Recall at 1 (R@1), which are described in detail in the supplementary materials. We compare our~{\SystemName}~ method, with both distillation and non-distillation baselines. For a fair comparison between different methods, the same low resolution is used at $128 \times 128$ pixels as mentioned above is used throughout the experiments.

\textbf{Non-distillation methods}. We reproduce two highest-score deepfake detection benchmark methods ~\cite{faceforensicsbenchmark}: 1) the method proposed by R\"ossler \emph{et al.} ~\cite{rossler2019faceforensics++}, which used Xception model, and 2) the approach by Dogonadze \emph{et al.} \cite{dogonadze2020deep}, which employed Inception ResNet V1 pre-trained on the VGGFace2 dataset \cite{cao2018vggface2}. These are the two best performing publicly available deepfake detection methods\footnote{\url{http://kaldir.vc.in.tum.de/faceforensics\_benchmark/}}. Additionally, we use the $F^{3}Net$, which is a frequency-based deepfake detection introduced by Quian \emph{et al.} \cite{qian2020thinking} for evaluation. The $F^{3}Net$ is deployed on two streams of XceptionNet as described in the paper. Finally, ResNet50 \cite{he2016deep} is also included as a baseline to compare with distillation methods. 

\textbf{Distillation baseline methods}.  As there has not been much research that deploys KD for deepfake detection, we further integrate other three well-known distillation architectures in the ResNet50 backbone to perform comparisons, including:  FitNet \cite{romero2014fitnets}, Attention Transfer \cite{zagoruyko2016paying} (AT) and  Non-local \cite{wang2018non} (NL). Each of these methods is fine-tuned on the validation set to achieve its best performance.

First, comparing ours with the non-distillation baselines, we can observe that our method improves the detection accuracy from $1\%$ to $6\%$ across all five datasets for both compression data types. On average, our approach outperforms the other three distillation methods, and is superior on the highest compressed (c40) datasets. The model with FitNet loss, though it has a small improvement, does not have competitive results due to retaining insufficient frequency information. The attention module and non-local module also provide compelling results. However, they do not surpass our methods because of the lower attention dimension and frequency information shortage.

\begin{table}[t!]
\centering
\resizebox{.48\textwidth}{!}{%
    \begin{tabular}{@{\hskip1pt}c  @{\hskip0pt}l  c c  c c@{\hskip2px}}
    \Xhline{2\arrayrulewidth}
    \multirow{2}{*}{Datasets}       & \multirow{2}{*}{\makecell{  \textrm{ }\textrm{ }\textrm{ }Models}}   & \multicolumn{2}{c}{\makecell{Medium comp-\\ression (c23)}}   & \multicolumn{2}{c}{\makecell{High comp-\\ression (c40)}}  \\ 
    \cmidrule(lr){3-4} \cmidrule(lr){5-6}
                                    &                              & ACC     & R@1   & ACC    & R@1 \\
    \Xhline{2\arrayrulewidth}
    \multirow{8}{*}{\rotatebox[origin=c]{90}{\centering NeuralTextures}} & R\"ossler \emph{et al.}  & 76.36  & 57.24 & 56.75  & 51.88   \Tstrut\\
                                    & Dogonadze \emph{et al.}   & 78.03  & 77.13  & 61.12 & 48.01  \\
                                    & $F^{3}Net$    & 77.91  & 77.39  & 61.95 & 32.35   \\
                                    & ResNet50    & 86.25  & 82.75  & 60.27 & 53.06   \\
                                    \cdashlinelr{2-6}
                                    & FitNet - ResNet50     & 86.26 & 84.83  & 66.01  & 57.28   \\
                                    & AT - ResNet50    & 85.21  & 84.99  & 62.61  & 43.50   \\
                                    & NL - ResNet50     & 88.26  & 86.95  & 65.65  & 46.82   \\ 
                                    &\SystemName~- ResNet50 (\emph{ours})  & \textbf{88.48}  & \textbf{87.53}  & \textbf{68.53}  & \textbf{58.42}       \Bstrut\\
    \hline
    \multirow{8}{*}{\rotatebox[origin=c]{90}{\centering DeepFakes}}      & R\"ossler \emph{et al.}   & 97.42 & 96.96  & 92.43  & 82.39  \Tstrut\\
                                    & Dogonadze \emph{et al.}    & 94.67  & 94.39  & 93.97  & 93.52   \\
                                    & $F^{3}Net$    & 96.26  & 95.84  & 93.06 & 93.00   \\
                                    & ResNet50   & 96.34 & 95.90  & 92.89 & 91.18 \\
                                    \cdashlinelr{2-6}
                                    & FitNet - ResNet50    & 97.28 & 97.78  & 93.68  & 93.34   \\
                                    & AT - ResNet50    & 97.37 & \textbf{98.72}  & 95.11  & 94.35  \\
                                    & NL - ResNet50    & 98.42  & 98.21  & 93.09   & 94.35  \\ 
                                    & \SystemName~- ResNet50 (\emph{ours})  & \textbf{98.67}  &  98.09  & \textbf{95.50}  & \textbf{94.59}      \Bstrut\\
    \hline
    \multirow{8}{*}{\rotatebox[origin=c]{90}{\centering Face2Face}}      & R\"ossler \emph{et al.}  & 91.83  & 91.02  & 80.21   & 77.42 \Tstrut\\
                                    & Dogonadze \emph{et al.}    & 89.34  & 88.73  & 83.44 & 81.00   \\
                                    & $F^{3}Net$    & 95.52  & 95.40  & 81.48 & 79.31   \\
                                    & ResNet50   & 95.60 & 94.77  & 83.94  & 79.88  \\
                                    \cdashlinelr{2-6}
                                    & FitNet - ResNet50    & 95.91  & 96.16  & 83.48  & 78.99  \\
                                    & AT - ResNet50    & 96.80 & 96.84  & 83.55  & 78.72  \\
                                    & NL - ResNet50    & 96.44  & 96.64  & 83.69  & 82.04  \\ 
                                    & \SystemName~- ResNet50 (\emph{ours})  & \textbf{96.82}  & \textbf{97.14}  & \textbf{85.42}   & \textbf{83.54}        \Bstrut\\
    \hline
    \multirow{8}{*}{\rotatebox[origin=c]{90}{\centering FaceSwap}}       & R\"ossler \emph{et al.}  & 95.49  & 95.36  & 88.09   & 87.67  \Tstrut\\
                                    & Dogonadze \emph{et al.}   & 93.33  & 92.78  & 90.02 & 89.10   \\
                                    & $F^{3}Net$    & 95.74  & 95.65  & 89.58 & 88.90   \\
                                    & ResNet50   & 92.46  & 90.85   & 88.91  & 86.52  \\
                                    \cdashlinelr{2-6}
                                    & FitNet - ResNet50    & 97.29  & 96.29  & 89.16    & 90.13 \\
                                    & AT - ResNet50    & 97.66 &  97.27  & 89.75  & 90.41  \\
                                    & NL - ResNet50    & 97.34  & 96.95  & 91.86    & 90.78 \\ 
                                    & \SystemName~- ResNet50 (\emph{ours})  & \textbf{97.85}   & \textbf{97.34}  & \textbf{92.49}   & \textbf{92.13} \Bstrut \\
    \hline
    \multirow{8}{*}{\rotatebox[origin=c]{90}{\centering FaceShifter}}    & R\"ossler \emph{et al.}  & 93.04 & 93.16  & 89.20   & 87.12 \Tstrut\\
                                    & Dogonadze \emph{et al.}   & 89.80  & 89.36  & 82.03 & 79.96  \\
                                    & $F^{3}Net$    & 95.10  & 95.02  & 89.13 & 88.69   \\
                                    & ResNet50    & 94.89  & 93.88  & 89.56   & 88.48 \\
                                    \cdashlinelr{2-6}
                                    & FitNet - ResNet50    & \textbf{96.63}   & 95.95  & 90.16  & 89.36  \\
                                    & AT - ResNet50   & 96.32 & \textbf{96.76}   & 88.28   & 89.45  \\
                                    & NL - ResNet50    & 96.24   & 95.28  & 90.04 & 87.71  \\ 
                                    & \SystemName~- ResNet50 (\emph{ours})  & 96.60  & 95.84  & \textbf{91.64}     & \textbf{90.27} \\
    \Xhline{2\arrayrulewidth}
    \end{tabular}%
}
\caption{Experimental results of our proposed method and other seven different baseline approaches on five different deepfake datasets. The best results are highlighted in bold.} 
\label{table:results_main}
\end{table}

\section{Ablation Study and Discussions}
\label{ablation}
\textbf{Effects of Attention Modules}.
We investigate the quantitative impact of the frequency attention and multi-view attention on the final performance. In the past, the NeuralTextures (NT) dataset has shown to be the most difficult to differentiate by both human eyes and DNN \cite{rossler2019faceforensics++}. Hence, we conduct our ablation study on the c40 highly NT compressed dataset. The results are presented in Table \ref{table:results_e_attention}. We can observe that frequency attention improves about $6.76\%$ of the accuracy. Multi-view attention with contrastive loss provides a slightly better result than that of without contrastive at $68.14\%$ and $67.01\%$, respectively. Finally, combining the frequency attention and multi-view attention distillation with contrastive loss significantly improves the accuracy up to $68.53\%$. The results of our ablation study demonstrate that each proposed attention distiller has a different contribution to the student's ability to mimic the teacher, and they are compatible when integrated together to achieve the best performance.

\begin{table}[h!]
\centering
\resizebox{.37\textwidth}{!}{%
\begin{tabular}{l | c }
\Xhline{2\arrayrulewidth}
Model            & ACC($\%$)       \\
\Xhline{2\arrayrulewidth}
ResNet (baseline)     & 60.27    \\
Our ResNet (FR)  & {67.03}           \\
Our ResNet (MV w/o contrastive)  & {67.01}   \\
Our ResNet (MV w/ contrastive)  & {68.14}         \\
Our ResNet (FR+MV)  & \textbf{68.53}     \\
\Xhline{2\arrayrulewidth}
\end{tabular}%
}
\caption{The effect of each single attention module on the final results experimented on NeuralTextures dataset.} 
\label{table:results_e_attention}
\end{table}

\textbf{Sensitivity of attention weights ($\alpha$ and $\beta$)}.
We conduct an experiment on the sensitivity of the frequency attention weight $\alpha$  and multi-view attention weight $\beta$ on the five different datasets. The detailed results are presented in the supplementary materials.
The result shows that by changing the value of $\alpha$ and $\beta$, the performance of our method continuously outperform the baseline results, indicating that our approach is less sensitive to both $\alpha$ and $\beta$. 

\textbf{Experiment with other backbones}.
Table \ref{table:results_e_resnet_18_34} shows the results with three other backbones: ResNet18 and ResNet34 \cite{he2016deep}, and EfficientNet-B0 \cite{tan2019efficientnet}. 
We set up the hyper-parameters of the four DNNs as the same for ResNet50, except $\gamma_{\mathcal{FR}}$ is changed to $1e^{-3}$ for EfficientNet-B0. Our distilled model improves the detection accuracy of all five datasets in different compression quality, up to $7\%$, $5.8\%$ $7.1\%$ with ResNet18, ResNet34, and EfficientNet-B0 backbone compared to their baselines, respectively.

\begin{table}[t!]
\centering
\resizebox{.48\textwidth}{!}{%
\begin{tabular}{@{\hskip4pt}l l  @{\hskip4pt}r  l  @{\hskip4pt}r l @{\hskip4pt}r l @{\hskip4pt}r l @{\hskip4pt}r}
\Xhline{2\arrayrulewidth}
& \multicolumn{2}{c}{\emph{\small{NT}}}  & \multicolumn{2}{c}{\emph{\small{DeepFakes}}}  & \multicolumn{2}{c}{\emph{\small{Face2Face}}}  & \multicolumn{2}{c}{\emph{\small{FaceSwap}}}  & \multicolumn{2}{c}{\emph{\small{FaceShifter}}}  \\
\cmidrule(lr){2-3} \cmidrule(lr){4-5} \cmidrule(lr){2-3} \cmidrule(lr){6-7} \cmidrule(lr){8-9} \cmidrule(lr){10-11}  \cmidrule(lr){4-5}
& c23 & c40 & c23 & c40 & c23 & c40 & c23 & c40 & c23 & c40 \\
\Xhline{2\arrayrulewidth}
\multicolumn{11}{c}{\textbf{ResNet18}}    \Tstrut \\
\Xhline{1\arrayrulewidth}
\emph{Baseline}&81.8    &67.3    &97.5    &89.2    &94.2    &85.0    &90.2    &84.5    &93.2    &89.2 \Tstrut\\
\emph{\SystemName} &84.3    &67.5    &97.7    &94.7    &95.7    &85.3    &96.0    &91.5    &97.0    &92.2 \Bstrut\\
\Xhline{2\arrayrulewidth}
\multicolumn{11}{c}{\textbf{ResNet34}}   \Tstrut  \\
\Xhline{1\arrayrulewidth}
\emph{Baseline}  &82.6    &58.4    &92.0    &93.4    &94.2    &83.2    &92.4    &88.6    &95.6    &89.3 \Tstrut\\
\emph{\SystemName} &84.3    &63.5    &97.8    &94.6    &94.9    &83.4    &96.8    &90.6    &97.8    &91.5 \Bstrut\\
\Xhline{2\arrayrulewidth}
\multicolumn{11}{c}{\textbf{EfficientNet-B0}}   \Tstrut  \\
\Xhline{1\arrayrulewidth}
\emph{Baseline} &81.2    &60.5    &96.5    &90.0    &94.1    &77.4    &92.6    &83.4    &93.8    &84.0 \Tstrut\\
\emph{\SystemName} &83.5    &67.6    &97.5    &92.5    &96.7    &80.3    &95.3    &87.5    &95.1    &85.2 \Bstrut\\
\Xhline{2\arrayrulewidth}
\end{tabular}%
}
\caption{Classification accuracy $(\%)$ of ResNet18, ResNet34 and EfficientNet-B0 baseline and their integration with our ADD training framework.} 
\label{table:results_e_resnet_18_34}
\end{table}

\textbf{Grad-CAM }\cite{selvaraju2017grad}. Using Grad-CAM, we provide visual explanations regarding the merits of training a LQ deepfake detector with our \SystemName~framework. The gallery of Grad-CAM visualization is included in the supplementary material. First, our ~\SystemName~ is able to correct the facial artifacts' attention of the LQ detector to resemble its teacher trained on raw datasets. Second, the ~\SystemName~ vigorously instructs the student model to neglect the background noises and activate the facial areas as its teacher does  when encountering facial images in complex backgrounds. Meanwhile, the baseline model which is solely trained on LQ datasets steadily makes wrong predictions with high confidence by activating non-facial areas and is deceived by complex backgrounds.

\section{Conclusion}
In this paper, we proposed a novel Attention-based Deepfake detection Distillations (ADD), exploring frequency attention distillation and multi-view attention distillation in a KD framework to detect highly compressed deepfakes. The frequency attention helps the student to retrieve and focus more on high-frequency components from the teacher. The multi-view attention, inspired by Sliced Wasserstein distance, pushes the student's output tensor distribution toward the teacher's, maintaining correlated pixel features between tensor elements from multiple views (slices). Our experiments demonstrate that our proposed method is highly effective and achieves competitive results in most cases when detecting extremely challenging highly compressed challenging LQ deepfakes. 
Our code is available here\footnote{\url{https://github.com/Leminhbinh0209/ADD.git}}.

\section{Acknowledgments}
This work was partly supported by Institute of Information \& communications Technology Planning \& Evaluation (IITP) grant funded by the Korea government (MSIT) (No.2019-0-00421, AI Graduate School Support Program (Sungkyunkwan University)), (No. 2019-0-01343, Regional strategic industry convergence security core talent training business) and the Basic Science Research Program through National Research Foundation of Korea (NRF) grant funded by Korea government MSIT (No. 2020R1C1C1006004). Also, this research was partly supported by IITP grant funded by the Korea government MSIT (No. 2021-0-00017, Original Technology Development of Artificial Intelligence Industry) and was partly supported by the Korea government MSIT, under the High-Potential Individuals Global Training Program (2020-0-01550) supervised by the IITP.

{
\small
\bibliography{aaai22.bib}
}
\begin{appendices}
\clearpage
\nocitesec{*}
\begin{alphasection}
\begin{LARGE}  
\textbf{Supplementary materials\\}   
\end{LARGE}

\section{A. Multi-View Attention Algorithm}
Algorithm {\ref{alg:multiview_attention}} presents the pseudo-code for multi-view attention distillation between two corresponding backbone features using the Sliced Wasserstein distance (SWD) from the student and teacher models. For implicity, we formulate how each single projection $\theta_i$ contributes to the total SWD. However, in practice, $K$ uniform vectors $\theta_{i}$ in $\mathcal{S}^{d-1}$ can be sampled simultaneously by deep learning libraries, \textit{e.g.}  TensorFlow or PyTorch, and the projection operation or binning can be vertorized.

\begin{algorithm*}[t!]
  \caption{Multi-view attention distillation using the Sliced Wasserstein distance (SWD)}
  \label{alg:multiview_attention}
  \begin{algorithmic}[1]
    \Require Two backbone features $\mathcal{A}_{\mathcal{S}}$ and $\mathcal{A}_{\mathcal{T}}$, which are the respective low-quality compressed and raw images from the student and teacher networks, the number of random projections $K$, and the number of bins $g$.
      \State $P_{\mathcal{A}_{\mathcal{S}}}$,\tab $P_{\mathcal{A} _{\mathcal{T}}}$ \tab $\gets$ \tab $\frac{\mathcal{A}_{\mathcal{S}}^{\circ 2} }{\Vert \mathcal{A}_{\mathcal{S}} \Vert_F^2}$,\tab $\frac{\mathcal{A}_{\mathcal{T}}^{\circ 2} }{\Vert \mathcal{A}_{\mathcal{T}} \Vert_F^2}$ \Comment{normalize $\mathcal{A}_{\mathcal{S}}$ and $\mathcal{A} _{\mathcal{T}}$}
      \State $SWD \gets 0$ \Comment{initialize SWD to zero} 
      \For{iteration $i \gets 1$ to $K$}
        \State $\theta_i$\tab $\gets$\tab $\mathcal{U}(\mathcal{S}^{d-1})$\Comment{uniformly sample $\theta_i$ from unit sphere in $\mathbb{R}^3$}
        
        \State $u_{\mathcal{S}}^{\theta_i}$,\tab $u_{\mathcal{T}}^{\theta_i}$\tab $\gets$ \tab $\mathcal{R}_{\theta_i}P_{\mathcal{A} _{\mathcal{S}}}$,\tab $\mathcal{R}_{\theta_i}P_{\mathcal{A} _{\mathcal{T}}}$ \Comment{project $P_{\mathcal{A}_{\mathcal{S}}}$ and $P_{\mathcal{A} _{\mathcal{T}}}$ on $\theta_i$}

        \State ${u}_{\mathcal{S}, \tau_1}^{\theta_i}$,\tab ${u}_{\mathcal{T}, \tau_2}^{\theta_i}$\tab $\gets $\tab  $Sort(u_{\mathcal{S}}^{\theta_i})$,\tab $Sort(u_{\mathcal{T}}^{\theta_i})$ \Comment{sort $u_{\mathcal{S}}^{\theta_i}$ and $u_{\mathcal{T}}^{\theta_i}$ in ascending order}
        
        \State $v_{\mathcal{S}}^{\theta_i}$,\tab $v_{\mathcal{T}}^{\theta_i}$\tab $\gets$\tab $\mathcal{G}({u}_{\mathcal{S}, \tau_1}^{\theta_i})$,\tab $\mathcal{G}({u}_{\mathcal{T}, \tau_2}^{\theta_i})$ \Comment{partition  ${u}_{\mathcal{S}, \tau_1}^{\theta_i}$ and   ${u}_{\mathcal{T}, \tau_2}^{\theta_i}$ into $g$ bins}
        
        \State $SWD \gets SWD + \sum_{j=1}^{g }\psi(v_{\mathcal{S}}^{\theta_i}[j], v_{\mathcal{T}}^{\theta_i}[j])$ \Comment{apply 1-D transportation cost with quadratic loss $\psi$}
      \EndFor\label{sampling}
      
      \State \textbf{return} $SWD $
  \end{algorithmic}
\end{algorithm*}
\label{supp:algorithm}

\section{B. Datasets}
\label{supp:dataset}
We describe the five different deepfake datasets used in our experiments:\
\begin{itemize}

    \item \textbf{NeuralTextures.} Facial reenactment is an application of \emph{Neural Textures}~\cite{thies2019deferred} technique that is used in video re-renderings. This approach includes learned feature maps stored on top of 3D mesh proxies, called neural textures, and a deferred neural renderer. The NeuralTextures datasets used in our experiment includes facial modifications of mouth regions, while the other face regions remain the same.\\
    
    \item \textbf{DeepFakes.} The DeepFakes dataset is generated using two autoencoders with a shared encoder, each of which is trained on the source and target faces, respectively. Fake faces are generated by decoding the source face's embedding representation with the target face's decoder. Note that DeepFakes, at the beginning, was a specific facial swapping technique, but is now referred to as AI-generated facial manipulation methods.  \\
    
    \item \textbf{Face2Face.} Face2Face~\cite{thies2016face2face} is a real-time facial reenactment approach, in which the target person's expression follows the source person's, while his/her identity is preserved. Particularly, the identity corresponding to the target face is recovered by a non-rigid model-based bundling approach on a set of key-frames that are manually selected in advance. The source face's expression coefficients are transferred to the target, while maintaining environment lighting as well as target background. \\
    
    \item \textbf{FaceSwap.} FaceSwap~\cite{faceswap} is a lightweight application that is built upon the graphic structures of source and target faces. A 3D model is designed to fit 68 facial landmarks extracted from the source face. Then, it projects the facial regions back to the target face by minimizing the pair-wise landmark errors and applies color correction in the final step. \\
    
    \item \textbf{FaceShifter.} FaceShifter~\cite{li2019faceshifter} is a two-stage face swapping framework. The first stage includes a encoder-based multi-level feature extractor used for a target face and a generator that has Adaptive Attentional Denormalization (AAD) layers. In particular, AAD is able to blend the identity and the features to a synthesized face. In the second stage, they developed a novel Heuristic Error Acknowledging Refinement Network in order to enhance facial occlusions.  \\
\end{itemize}

\section{C. Distillation Baseline Methods}
In our experiement, we actively integrate three well-known distillation losses into the teacher-student training framework for comparing with ours:

\begin{itemize}
    \item \textbf{FitNet} \cite{romero2014fitnets}. FitNet method proposed hints algorithms, in which the student's guided layers try to predict the outputs of teacher's hint layers. We apply this hint-based learning approach on the penultimate layer of the teacher and student. 
    \item \textbf{Attention Transfer} \cite{zagoruyko2016paying} (AT). Attention method transfers attention maps, which are obtained by summing up spatial values across the backbone features' channels from the teacher to the student. 
    \item \textbf{Non-local} \cite{wang2018non} (NL). Non-local module generates self-attention features from the student and teacher's backbone features. Subsequently, the student's self-attention tensors attempt to mimic the teacher's. 
\end{itemize}
\section{D. Evaluation Metrics}
\label{supp:metric}
The results in our experiments are evaluated based on the following metrics:
\begin{itemize}
    \item \textbf{Accuracy (ACC)}. ACC is widely used to evaluate a classifier's performance, and it calculates the proportion of samples whose true classes are predicted with the highest probability. ACC of a model $f_\theta$ tested on a test set of $N$ samples $\{(x_{1}, y_{1}),...,(x_{N}, y_{N})\}$ is formulated as follows:
    \begin{equation}
        ACC = \frac{\sum_{i=1}^{N} I\left(\argmax (f_{\theta}(x_{i})), y_{i}\right)}{N},
    \end{equation}
    where $I(\cdot,\cdot)$ is the  Kronecker delta function. 
    \item \textbf{Recall at $k$ ($R@k$)}. $R@k$ indicates the proportion of test samples with at least one observed sample from the same class in $k-$nearest neighbors determined in a particular feature space. A small $R@k$ implies small intra-class variation, which usually leads to better accuracy. $R@k$ is formulated as follows:
    \small{
    \begin{equation}
        R@k = \left(1 - \frac{\sum_{i=1}^{N} I\left(\sum_{j=1}^{k} I\left (neighbor_{x_{i}}[j] , y_{i}\right ) , 0 \right )}{N} \right),
    \end{equation}
    }
    where $neighbor_{x_{i}}[j]$ is the label of the $j^{th}$ nearest neighbor of $x_i$, and  $I(\cdot,\cdot)$ is the  Kronecker delta function. We use the Euclidean distance to measure the distances between queried and referenced samples whose features are the penultimate layer's outputs, and we adopt $R@1$ which considers the first nearest neighbor of a test sample.
\end{itemize}

\section{E. Hyperparameters' Sensitivity}
\label{supp:sensitivity}

\begin{figure*}[t!]
\centering
\includegraphics[width=5.2in]{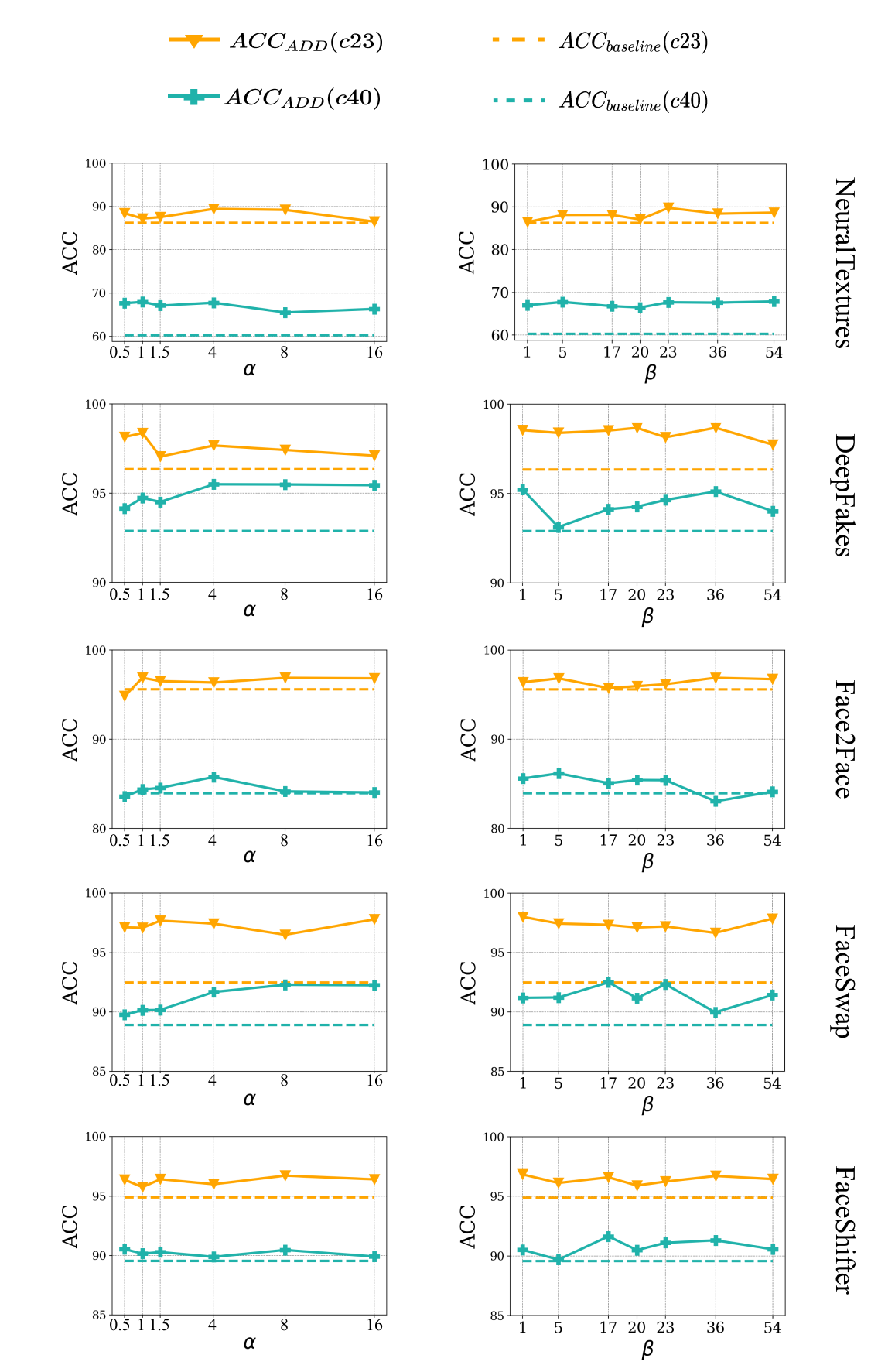}
\caption{Ablation study on attention weights $\alpha$ (left column) and $\beta$ (right column). The blue and orange lines indicate results with  medium compression (c23) and high compression (c40) data, respectively. The solid and dashed lines represent results of our ~\SystemName~- ResNet50 student model and the ResNet50 baseline model, respectively.} 
\label{fig:alpha_beta_val}
\end{figure*}
We examine the sensitivity of the different choices of attention weight hyperparameters, $\alpha$ and $\beta$. We experiment and show that our approach (solid lines) is less sensitive to $\alpha$ and $\beta$, and almost always outperforms the ResNet50 baselines (dashed lines) across all datasets, as shown in Fig. \ref{fig:alpha_beta_val}.
In this work, we fix $\alpha=1$  and fine-tune $\beta$, but they can be further tuned to optimize the performance.

\section{F. Gallery of Grad-CAM for the Teacher, Student and Baseline Model}
\label{supp:gradcam}
Grad-CAM~\cite{selvaraju2017grad} is generated by back-propagating the gradients of the highest-probability class to a preceding convolutional layer's output, producing a weighted combination with that output. The localization heat map visually indicates a particular contribution of each region to the final class prediction. The strong activation regions, which are represented in red color, are resulted from positive layer's outputs and higher value of the gradients, whereas negative pixels or low gradients produce less activation regions are indicated in blue color. In our experiments, we utilized ResNet50 pre-trained on raw data as the teacher, our~\SystemName~- ResNet50 trained on low-quality compressed data (c40) as the student, and ResNet50 trained on the c40 data alone without any KD method as a baseline for comparison. We provide visual explanations regarding the two benefits of training a low-quality deepfake detector with our~\SystemName~framework as follows: 
\begin{itemize}
    \item \textbf{Correctly identifying facial activation regions}. Despite being trained on low-quality compressed images, the ResNet50 baseline still makes wrong predictions with high confidence as shown in the second column of Fig. \ref{fig:gradcam_face}. The areas pointed by red arrows indicate the baselines' activation regions, which are different from the teacher's, indicated by green arrows in the third column. After training with our~\SystemName~framework, the student is able to produce correct predictions that are strongly correlated and can be further visualized by its activation regions which are closely following the teacher's. 
    
    \item \textbf{Resolving background confusion}. Figure \ref{fig:gradcam_bg} presents the Grad-CAMs for the fake class of five different datasets, the selected samples of which have complex backgrounds. We can observe that when the teacher is trained with raw data, it produces predictions (around 1.00) with nearly perfect confidence for the fake class. When encountered highly compressed images with complex backgrounds, the ResNet50 baseline model makes wrong predictions, not activating the facial regions but the background (red arrows in Fig. \ref{fig:gradcam_bg}). On the other hand, our~\SystemName~- ResNet50 model is also trained on low-quality compressed data, but our approach accumulates more distilled knowledge from the teacher and it is able to correctly identify and emphasize the actual facial areas, as its teacher similarly does with raw images (green arrows in Fig. \ref{fig:gradcam_bg}). 
    
    As shown in Fig.~\ref{fig:gradcam_bg}, we can demonstrate and conclude that when a low-quality compressed image has a complex background, it is easy for a conventional learning model trained without additional information (the second column) to become more vulnerable, making incorrect predictions with high confidence, activating background regions.  Meanwhile, with our KD framework, the student is trained under the guidance of its teacher that can focus more on the facial areas, and effectively eliminate the background effects from the real and fake faces. 
    
    Interestingly, as shown in Fig.~\ref{fig:gradcam_bg}, the baseline classifier can possibly be susceptible to and exploited by adversarial attacks, which typically add small amount of noise to the image. Therefore, one can explore the possibility of performing adversarial attacks on the compressed images by adding small amount of noise to or disordering the background. These changes resulting from adversarial attacks would be difficult to be distinguished by humans and can easily deceive the output of the baseline classifier. Therefore, it would be interesting to explore adversarial attacks for complex, compressed images and videos. Also, as a defense mechanism, we can consider KD as a framework with our novel attention mechanism for future work to better detect face regions and increase robustness against complex background noise as shown in Fig.~\ref{fig:gradcam_bg}.   

\end{itemize}

\begin{figure*}[h]
\centering
\includegraphics[width=5.6in]{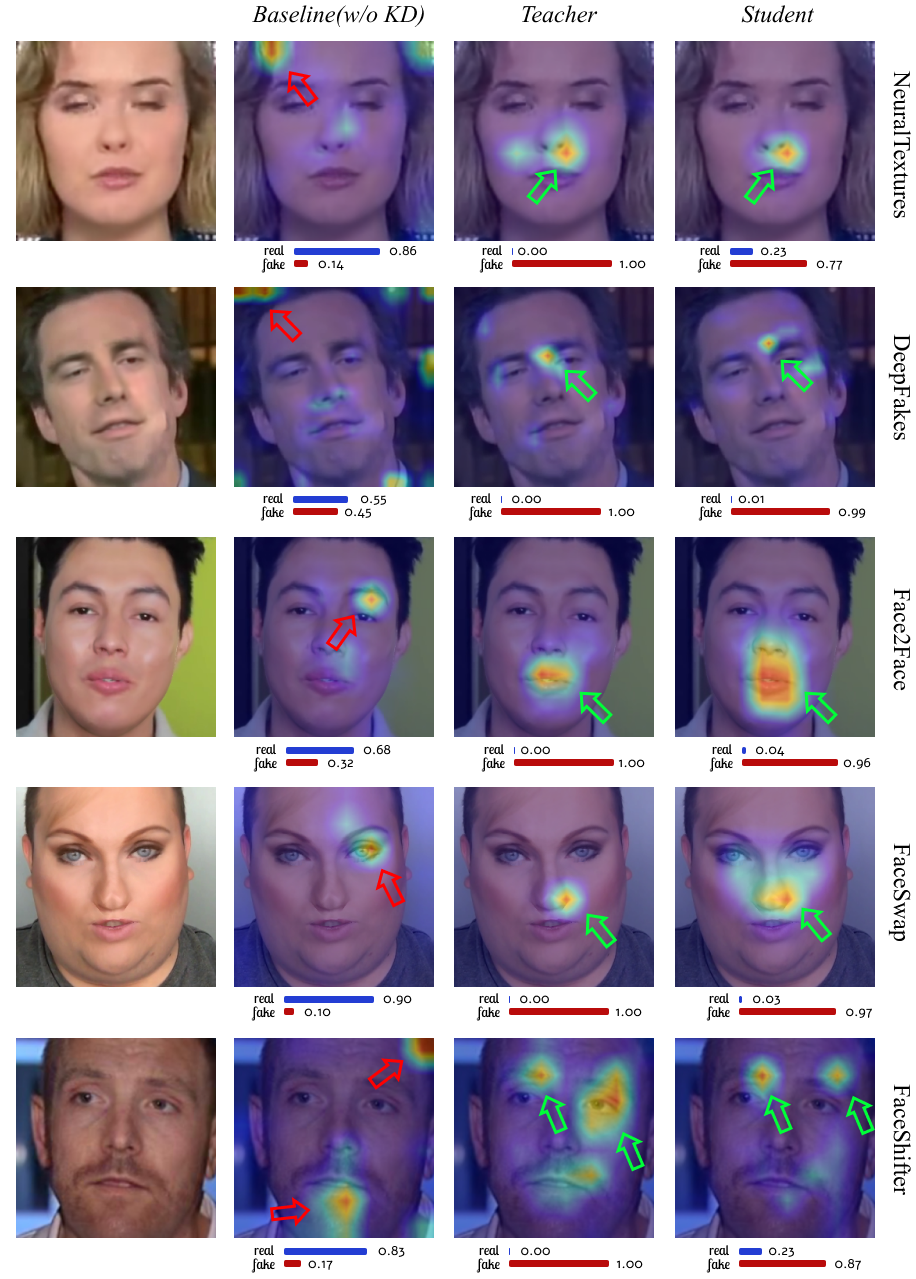}
\caption{Grad-CAM outputs of deepfake images from NeuralTextures, DeepFakes, Face2Face, FaceSwap and FaceShifter dataset. The strongest activation regions are highlighted in red, where the blue indicates the more silent ones. The red arrows indicate the activation regions that lead to wrong predictions of ResNet50 baseline trained on highly compressed datasets without our KD framework. Green arrows in the fourth column indicate the facial activation regions of the student, which almost match with the teacher's in the third columns, resulting from distilling knowledge from the teacher effectively by our~\SystemName~framework. \emph{(Figure is best viewed in color)}.}
\label{fig:gradcam_face}
\end{figure*} 

\begin{figure*}[h]
\centering
\includegraphics[width=5.6in]{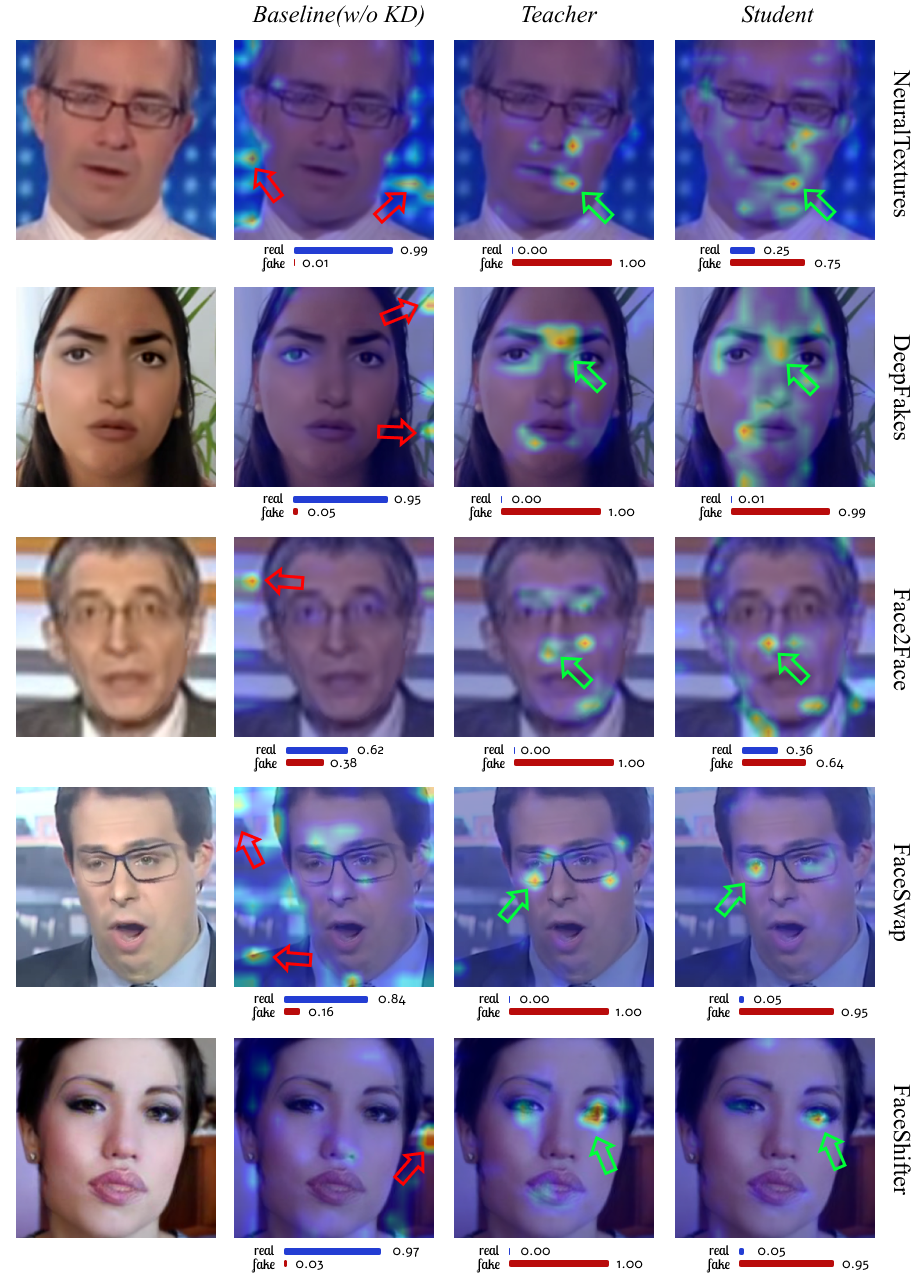}
\caption{Class activation maps of deepfake images from NeuralTextures, DeepFakes, Face2Face, FaceSwap and FaceShifter dataset. The green arrows shown the correct facial activation regions that the student successfully mimic its teacher. The red arrows indicate non-facial areas that ResNet50 baseline relies on to make the wrong predictions with high confidence when trained on highly compressed datasets. \emph{(Figure is best viewed in color)}.}
\label{fig:gradcam_bg}
\end{figure*}

\end{alphasection}
\end{appendices}
\end{document}